\documentclass[lettersize,journal]{IEEEtran}

\IEEEoverridecommandlockouts

\usepackage[
	citestyle=ieee, 
    bibstyle=ieee,
    style=numeric-comp,
    sorting=nty, 
    ]{biblatex}
\DeclareNameAlias{author}{family-given}
\usepackage{float}
\usepackage{amsmath,amssymb,amsfonts}
\usepackage{algorithmic}
\usepackage{caption}
\usepackage{graphicx}
\usepackage{textcomp}
\usepackage{xcolor}
\def\BibTeX{{\rm B\kern-.05em{\sc i\kern-.025em b}\kern-.08em
    T\kern-.1667em\lower.7ex\hbox{E}\kern-.125emX}}
\begin{document}

\title{Deep Learning on Graphs for Mobile Network Topology Generation}

\author{
    \IEEEauthorblockN{
        Felix Nannesson Meli\IEEEauthorrefmark{1}, Johan Tell\IEEEauthorrefmark{1}, Shirwan Piroti\IEEEauthorrefmark{2},  Tahar Zanouda\IEEEauthorrefmark{2}, Elias Jarlebring\IEEEauthorrefmark{1}
    } \\
    \IEEEauthorblockA{\IEEEauthorrefmark{1}KTH Royal Institute of Technology}, \IEEEauthorblockA{\IEEEauthorrefmark{2}Ericsson.} \\
    \texttt{\{felixnm, johantel, eliasj\}@kth.se}
    
\texttt{\{shirwan.piroti, tahar.zanouda\}@ericsson.com}\\
}

\maketitle

\begin{abstract}
Mobile networks consist of interconnected radio nodes strategically positioned across various geographical regions to provide connectivity services. The set of relations between these radio nodes, referred to as the \emph{mobile network topology}, is vital in the construction of the networking infrastructure. Typically, the connections between radio nodes and their associated cells are defined by software features that establish mobility relations (referred to as \emph{edges} in this paper) within the mobile network graph through heuristic methods. Although these approaches are efficient, they encounter significant limitations, particularly since edges can only be established prior to the installation of physical hardware.

In this work, we use graph-based deep learning methods to determine mobility relations (edges), trained on radio node configuration data and reliable mobility relations set by Automatic Neighbor Relations (ANR) in stable networks. This paper focuses on measuring the accuracy and precision of different graph-based deep learning approaches applied to real-world mobile networks. We evaluated two deep learning models. Our comprehensive experiments on Telecom datasets obtained from operational Telecom Networks demonstrate the effectiveness of the graph neural network (GNN)  model and multilayer perceptron. Our evaluation showed that considering graph structure improves results, which motivates the use of GNNs. Additionally, we investigated the use of heuristics to reduce the training time based on the distance between radio nodes to eliminate irrelevant cases. Our investigation showed that the use of these heuristics improved precision and accuracy considerably.
\end{abstract}


\section{Introduction}
Mobile networks are the backbone of modern communication infrastructure. These networks are constantly evolving and continuously expanding, especially as the networks have become more dense along with the introduction of 5G. These networks consist of connections between different radio nodes (base stations) that represent the network topology. Mobile networks consist of a set of radio nodes that are distributed in a geographical region to provide connectivity services. Each radio node spans a set of cells. Each radio node provides connectivity coverage to end-user equipment (UE). UEs are devices with connectivity capabilities such as smart phones, watches or smart cars. One of the key functionalities in a mobile network is the concept of a handover. A handover is the transfer of connection from a radio node to another. In 5G/NR and 4G/LTE networks, handovers are made possible by maintaining a fixed cell relation list, as described in  
\cite{Cloud_based_link_prediction}. This type of handover procedure provides possible neighboring radio nodes that a UE connects to when moving from one coverage area to another.

The handover relations between cells are typically determined by Software features named Automatic Neighbor Relations (ANR) \cite{dahlen11}. Traditionally, the handover relations are generated based on domain knowledge. However, traditional and statistical approaches are criticized for relying on domain knowledge and assumptions that can limit their applications to real-world scenarios. ANR uses network traffic through the movement of UEs in the network and handover statistics, among other data, to find the neighbours. Consequently, ANR can only be used after deployment of radio nodes or cells, and the process of deciding the initial cell relations might be slow. This work aims to find these relations for new radio nodes using information that is available before deployment.

In this paper, we investigate the use of graph-based deep learning to generate cell relations before deployment of new cells and radio nodes using attribute information from the network. To address this challenge, we represent a mobile network as a graph and formulate the problem as an inductive link prediction task. In order to encode the structure of the network graph we use an deep learning architecture based on graph neural networks (GNNs) \cite{gori2005new}.
This choice was inspired by previous work \cite{leatherman2023link} on inductive link prediction that has shown success in using neural networks and GNNs in a more general setting not specifically adapted for telecommunication networks. 
See Section~\ref{sec:relatedwork} for further discussion of related work.

The goal of this work is to automatically generate mobility relations between cells in a network before it is physically deployed. This is done by applying deep learning algorithms to predict the cell relations using the configuration data of the cells. 
\vspace{0.25cm}

\noindent

The main scientific contribution of this work is a new method for topology generation in telecom networks. To our knowledge, this is the first method for this type of telecom problem based on graph neural networks. The method includes data processing, architecture and model construction and quality quantification. The techniques to quantify the quality is constructed in a way that allows us to compare it with other methods.

\section{Related Work}\label{sec:relatedwork}
In recent years, there has been a growing body of research that focuses on applying graph-based deep learning for a wide range of applications in the telecommunications domain \cite{ShirwanGNN}. However, graph-based deep learning approaches have not been employed extensively to generate mobile network topology. Different methods have been employed to automate the process of generating network topology. Notably, Pina et al. \cite{Cloud_based_link_prediction}, Dahlén et al. \cite{dahlen11}, and Parodi et al. \cite{automatic_cell_list_def07} proposed different methods to automatically generate mobile network topology while reducing human intervention. 

Hao et al. \cite{DEAL} introduced \textit{DEAL}, this method to encode both attributes and information about the graph structure with multi-layer perceptrons (MLP) to get state-of-the-art results on both inductive and transductive link prediction. However, this model has the downside that it is too computationally expensive to run on large graphs. 

Samy et al. \cite{graph2Feat} proposed \textit{Graph2Feat} to train multilayer perceptrons to match the predictions of a variational graph autoencoder (VGAE) \cite{1611.07308} in a teacher-student framework. Leatherman \cite{leatherman2023link} improved previous work by setting some edges with a multilayer perceptron based on a score of importance and then using a VGAE to generate more edges.

The closest work we found was a method proposed by Pina et al. \cite{Cloud_based_link_prediction}, which addresses limitations of the ANR algorithm similar to the ones discussed in this paper by creating a cloud-based solution. The authors used a standard propagation model for cell coverage estimation to compute a world map for calculating neighbor relations. This approach, while exhibiting strong results, can be computationally heavy for large graphs. 

\section{Method}
\label{section:method}

The method that we propose consists of a choice of model and deep learning architecture, as we describe in Section~\ref{sec:model}. In order to fully describe the method and do proper model selection, we first define the problem (Section~\ref{sec:ran-graph}) and the data (Section~\ref{sec:Pre-processing}).

\subsection{RAN Graph}
\label{sec:ran-graph}
In this work, we let the mobile telecommunications network be represented as an undirected unweighted graph $G=(V,E)$, where cells correspond to the nodes (vertices) and their relations as edges.

Let $V=\{1, \cdots n\}$ be the node set where $n$ is the number of nodes, and $E=\{\{i,j\}, \cdots | i,j \in V\}$ be the edges. Additionally, define $e_i$ to be the set of all the edges of node $i$. 
With that, let $H_i=(V\setminus \{i\}, E\setminus e_i)$ be the graph with node $i$ removed and $\varphi: V \longrightarrow \mathbb{R}^k$ be a mapping from the edge set $V$ to the $k$-dimensional space $\mathbb{R}^k$. The elements of the mapped sets through $\varphi$ are called \textit{attributes}. 
\emph{The objective is to find the added edges to the current nodes 
given a new node to the graph $n+1$ with attributes $\varphi(n+1)$.} 
In other words, we wish to find a function $f(G, \varphi, \varphi(n+1)) = e$ such that $e$ represents the edges of node $n+1$.

\subsection{Data}\label{sec:Pre-processing}

The data for this study is collected from multiple Evolved Node Bs (eNBs) and Next Generation Node Bs (gNBs),which is collected and processed in the following way.

\subsubsection{Pre-processing and Feature Engineering} 

In this study, the data is collected from multiple eNBs
and gNBs in an operational network. When the network is operational, operators continuously track the behavior of the network using \textit{Performance Management (PM) data}. PM data is captured regularly across Radio nodes. \textit{Configuration Management (CM) data} characterizes network structure through a multitude of configuration parameters (e.g., frequency, technology, hardware products, etc.). We calculate Key Performance Indicators (KPIs) based on PM and CM data to determine stable networks.

To this end, the network data is processed to represent the network mathematically by a graph, where each cell in the telecom network is represented by a feature vector.

Finally, the data was normalized over the attributes of the graph using z-score normalization.

\subsubsection{Data Filtering}
Once the data is prepared, we preprocess the data for the baseline. The baseline consists of what we call a candidate algorithm, which takes a certain amount of nodes within a maximum distance to filter the nodes for those that are too far away while ensuring that candidates are still provided when there are not many nodes that lie in very close proximity to each other. For certain variations of configurations and data topology, this approach can be formulated as considering the 2D K-Nearest Neighbors applied only on the coordinate features (lat, long). The different configurations of $k$ (max number of candidates) and $m$ (max distance to candidate) can have varying effects on prediction performance. Thus, experiments were conducted to measure which configuration may be most reliable for the problem.

\subsubsection{Train \& Validation datasets}

During the different training phases, the data is masked so that the edges from the later phases are not included. This is done to prevent data leakage in the models.

The data was split into training \textit{($90\%$)}, validation \textit{($5\%$)}, and testing \textit{($5\%$)}. The reason for this split ratio is that it is close to the common $80/10/10$, but since not all the training data was used at the same time, and there was such a large amount of nodes available that $5\%$ validation and $5\%$ testing data is sufficient. To ensure this was the case, tests with an $80/10/10$ split were made, of which the resulting difference was negligible. 

\subsection{Model Definition \& Architecture}\label{sec:model}
The models developed in this study are classifiers for the inductive link prediction task. The baseline model consists of an MLP, while the main model is \textit{GNN-based}.

\subsubsection{Baseline model}

The model consists of an MLP where the node features of two possible related nodes are concatenated and passed through as input, with a single value between $0$ and $1$ as output, representing the probability of the nodes being related.  The high-level architecture is shown in figure~\ref{fig:MLP}. The mathematical definition of the system of \textit{MLP} is given by:
\vspace{-0.5cm}
    
\begin{align*}
&\text{MLP}(i,j) =\\
&\sigma(\bold{W}_3 \text{ReLU}(\bold{W}_2 \text{ReLU}(\bold{W}_1 \text{concat}(\bold{X}_i, \bold{X}_j) + \bold{b}_1)+\bold{b}_2)+\bold{b}_3).\\
\end{align*}
\noindent

\subsubsection{Main model}

The \textit{GNN-based model} consists of a single layer of SAGE-layers \cite{graphSAGE} taking the unmasked graph to encode the input features of the nodes into higher space feature embeddings. These embeddings also include information on the network topology through the averaging over edges used in the SAGE layer. These embeddings are then used as features for an MLP that takes the concatenated feature embeddings as input and outputs a single value between $0$ and $1$ representing the probability of the nodes being neighbors. 
Our benchmarks showed that using SAGE provided better results than the conventional graph convolutional layers leading us to use the former for the models. The model architecture is depicted in \ref{fig:GNN-MLP}.
The mathematical definition of the system of \textit{GNN-based model} is given by:

For $N$ number of nodes and $\bold{E} = \text{ReLU}(\text{SAGE}(\bold{X}, \bold{A}))$ we have 
\begin{align*}
    &\text{GNN}_2(i, j) = \\
    &\sigma(\bold{W}_3 \text{ReLU}(\bold{W}_2 \text{ReLU}(\bold{W}_1 \text{concat}(E_i, E_j) + \bold{b}_1)+\bold{b}_2)+\bold{b}_3) \\
\end{align*}
The quantities in the expression are feature matrix $X \in \mathbb{R}^{8\times N}$,
embedding matrix 
$E\in\mathbb{R}^{64\times N}$, 
adjacency matrix $A \in \mathbb{R}^{N\times N}$, 
weight matrices $W_1 \in \mathbb{R}^{64\times16}, W_2 \in \mathbb{R}^{64\times64}, W_3 \in \mathbb{R}^{1\times64}$ 
and bias vectors $b_1 \in \mathbb{R}^{64}, b_2 \in \mathbb{R}^{64}, b_3 \in \mathbb{R}^1$.

\begin{center}
\begin{figure}[h]
    \centering
    \includegraphics[width=\linewidth]{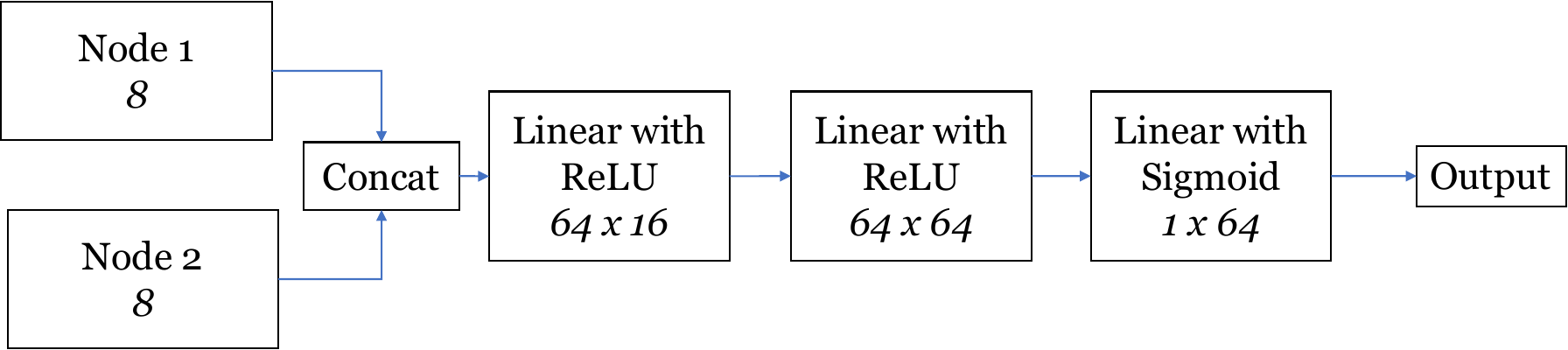}
    \caption{Block diagram of \textit{MLP}.}
    \label{fig:MLP}
    
\end{figure}
\begin{figure}[h]
    \centering
    \includegraphics[width=\linewidth]{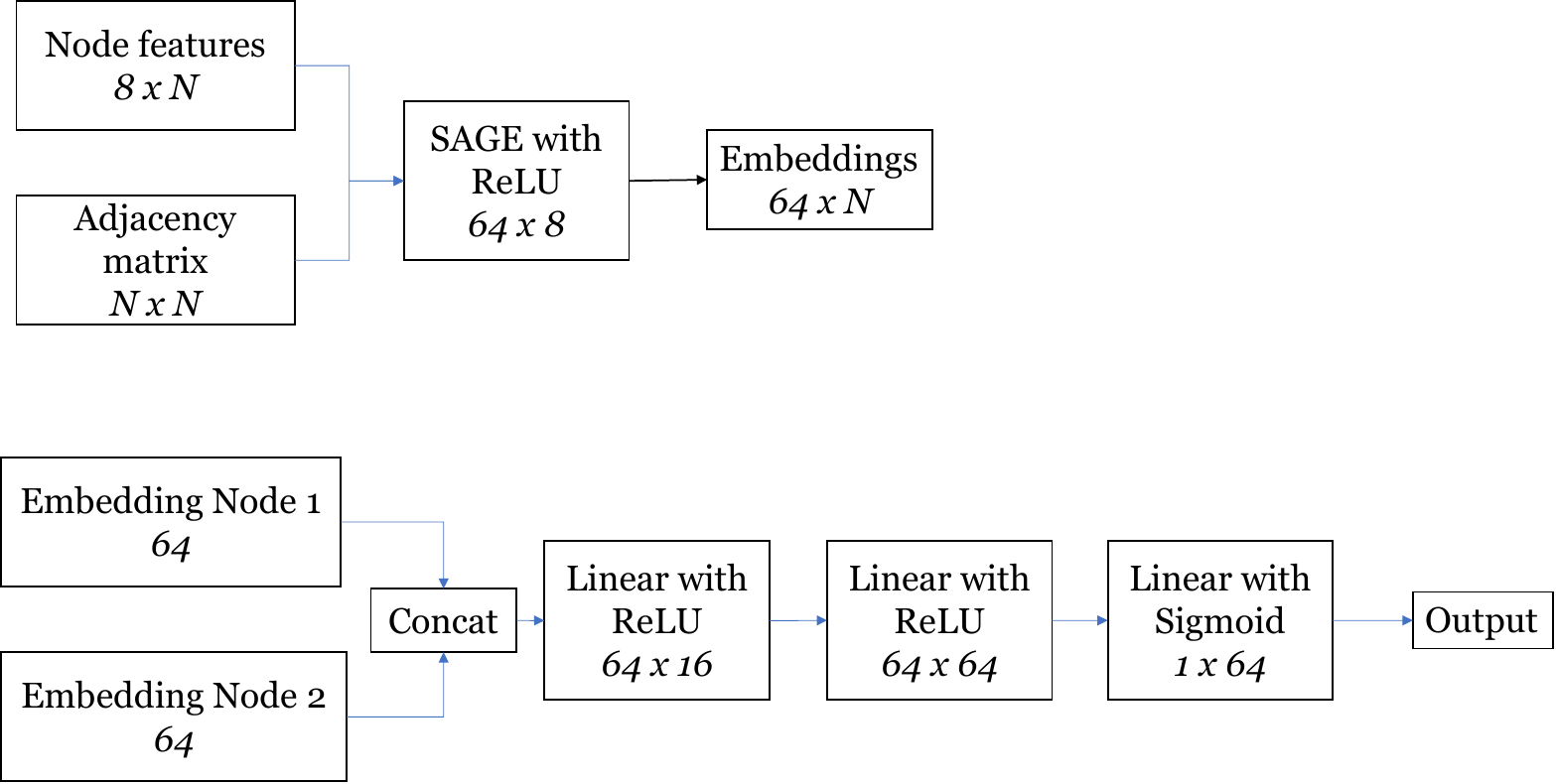}
    \caption{Block diagram of \textit{GNN}. The upper part of the diagram produces the feature embeddings for the nodes, which are then used for the sampled possible neighbors as input to the lower diagram as in the MLP (figure \ref{fig:MLP}).}
    \label{fig:GNN-MLP}
    
\end{figure}
\end{center}

\subsection{Discussion}

\begin{table}[H]
    \centering
    \begin{tabular}{c|c|c|c}
         Configuration & Precision  & Recall & Accuracy \\
         \hline
         $K$=300, $m$=450  & 0.1177 & \textbf{ 1.000} & \textbf{0.9999}\\
        $K$=40, $m$=450& \textbf{0.6592} & 0.7294 & 0.9990\\
        $K$=50, $m$=50  & 0.5647 & 0.2476  & 0.9972\\
        \hline
    \end{tabular}
    \caption{Results from the candidate algorithm on the validation data for different parameters. Note that $m=450$ is unreasonable large which means that it may be more feasible to use only the $K$ nearest nodes, as is done in this report.}
    \label{tab:cand-results}
\end{table}
\vspace{0.5cm}

\begin{table}[H]
\centering

\begin{tabular}{|c|c|c|c|c|}
\hline
 & ACC (\%) & Precision (\%) & Recall (\%) & AUC \\ \hline
MLP (baseline) & 94.36 & \textbf{93.16} & 95.73 & \textbf{0.99} \\ \hline
GNN & \textbf{94.70} & 92.73 & \textbf{97.00} & \textbf{0.99} \\ \hline
\end{tabular}

\caption{Results for validation where an equal number of non neighbours to neighbours was used}
\label{tab:validation1}

\begin{tabular}{|c|c|c|c|c|}
\hline
 & ACC (\%) & Precision (\%) & Recall (\%) & AUC \\ \hline
MLP (baseline) & 93.46 & 2.64 & 95.73 & \textbf{0.99} \\ \hline
GNN & \textbf{94.56} & \textbf{3.19} & \textbf{97.00} & \textbf{0.99} \\ \hline
\end{tabular}

\caption{Results for validation when using all neighbours and non neighbours}
\label{tab:validation2}

\begin{tabular}{|c|c|c|c|c|}
\hline
& ACC (\%) & Precision (\%) & Recall (\%) & AUC \\ \hline
MLP (baseline) & 19.58 & 12.43 & 95.73 & 0.67 \\ \hline
GNN & \textbf{23.45} & \textbf{13.10} & \textbf{97.00} & \textbf{0.7} \\ \hline
\end{tabular}

\caption{Results for validation on output from the candidate algorithm using settings $K=300, m=450$. Since the candidate algorithm has a recall of $1$, every neighbour is in this set. Note that the accuracy is how well the model can classify the output from the candidate algorithm}
\label{tab:validation3}
\end{table}

The results for the full unbalanced data indicate that the precision of the models is quite low with a maximum of 13.10 \% even with our candidate algorithm. We can also see that the recall is generally quite high, with a minimum of 72.26 \%. We can then conclude that our models generally produce many false positives while getting few false negatives. This can be attributed to the imbalance in our dataset and could potentially be remedied by training the model on another distribution of positives to negatives. Of course, with too many negative samples the problem of having most outputs default to zero may arise. Another way to remedy this is to weigh the loss for the positives and negatives differently or use another loss function entirely. 

Comparing the different models, we see that \textit{GNN} scores highest when considering accuracy and precision. It is not surprising that \textit{GNN} has higher results than \textit{MLP} since \textit{GNN} is in a sense just \textit{MLP} but with a SAGE layer on top. If a different balance between recall and precision is needed this could be acquired by shifting where the cutoff point is for what is considered to be a positive ($1$) or negative ($0$) output for \textit{MLP}.

The high scoring results of the candidate algorithm suggest that one of the most important features, if not the most important, is the geographical location of the cell. At the same time, we see that the performance of the candidate algorithm may be bounded below the deep learning models' performance, given the results of especially the second configuration ($K = 40$,$m=450$). We can observe this in the difference in performance of the configurations in \ref{tab:cand-results}.

By looking at the results from applying the candidate algorithm to the models in table~\ref{tab:validation3} and comparing it to the results in table~\ref{tab:validation2} it can be observed that on the unbalanced validation nodes the candidate algorithm helps to achieve considerably higher results for precision than just using the deep learning models. This might suggest that the candidate algorithm could act as a filter for producing a good split of the data for the deep learning models. Note that the total accuracy was reduced substantially, which most likely is due to having such low precision in the candidate algorithm for the specific settings used ($K=300$, $m=450$), leading to a highly imbalanced data distribution while being relatively limited in one of the most important features, geographical location. 

In practice it is important not to set too many relations because it would be computationally costly to run such a network, leading to high energy consumption and unnecessary computation. A reasonable approach would be to pick a maximum number of neighbours for every node. By then applying the models and picking the most confident predictions over a certain threshold, the distilled predictions might be more accurate and precise. Another way of doing this is to change the cutoff for neighbor prediction based on validation data in \textit{MLP} and \textit{GNN} and thus hopefully extracting higher quality predictions.

Another factor to take into account is if these results are useful. To clarify, in this problem formulation the aim is to replicate the relations that today are set by ANR, it is however difficult to know if there could be other equivalent network topologies that, while not being the ones currently set by ANR, could be just as, if not more effective. A reasonable assumption for this work is that the relations set by ANR are close to the optimal solution since it is based on real handover data. However, this assumption is not necessary to make, and a different perspective may even prove to be fruitful.

It is important to also consider the computational cost of implementing a solution using the models in practice, since the models may be running a large number of configurations making it advantageous to have higher computational efficiency. If the models' computational costs are too high they may be too expensive to use. For example, while \textit{MLP} might not have the same level of performance as \textit{GNN}, the complexity may be reduced for the computations, and it may prove more viable to use. It also does not require an existing graph structure to work.

When implementing these models in practice, it is crucial to respect that one significant feature absent from our dataset, the bearings of the cells, is indicated as important in Pina et al. \cite{Cloud_based_link_prediction}. Incorporating this data could potentially improve our predictions. Furthermore, another contributing factor is that the location data for the cells reflects the position of the computational hardware managing a group of cells rather than the actual position of the antenna, which may render some data points less reliable.

Another limiting factor is missing data; there was a substantial amount of nodes that lacked some feature(s) and it has not been explored how these nodes would affect the results. Limited computational resources also posed a challenge, making it difficult to train on all available data due to extended training times or to employ more resource-intensive models during the training phase.

Further work on this subject could involve the advanced models described in Samy, Kefato and Girdzijauskas\cite{graph2Feat} or Leatherman \cite{leatherman2023link}, which both use variational graph autoencoders to generate the structure of every edge at the same time. This would be computationally faster and potentially have better results. If possible one could also add other attributes, such as bearing and possibly population density around the cell - which could improve the results. The bearing may aid in differentiating cells close together but pointed away from each other, and population density could point to how dense the cellular network is in an area.
This information can indicate the appropriate number of cell relations.

An important idea that remains unexplored is utilizing is using the high scores of \textit{MLP} and \textit{GNN} on balanced data by first producing a balanced subset of the data through the candidate algorithm. We have already shown in this paper that it is possible to get around $50\%$ precision with the candidate algorithm (table \ref{tab:cand-results}), which would enable us to filter the data to get an almost $50/50$-split subset. This could help bridge the gap between the performance on the real-world versus ideal class ratios.

To adress the problem with assuming that ANR produces the unique optimal solution, one could instead construct graphs and measure graph similarity, match subgraphs within the resulting network or in other ways quantify the similiarity or dissimilarity in some sense, in order to more properly determine the efficacy of the models. This is especially important if generative models like VGAE is used, since there may be distributed "off by one" errors that still give an equivalent network topology.

\section{Conclusion}
In this paper, the proposed framework enables mobile
operators to, pre deployment, generate mobility relations in telecom networks. A solution method using graph-based deep learning model was proposed. The model generates relations by predicting the edge between two nodes in the graph. The model was evaluated on datasets from operational networks, and against another baseline, achieving high accuracy on the datasets, indicating that the solution is generalizable.

\printbibliography

\end{document}